\documentclass{article}
\usepackage{ijcai18}

\usepackage{times}
\usepackage{xcolor}
\usepackage{soul}
\usepackage[utf8]{inputenc}
\usepackage[small]{caption}

\usepackage{helvet}
\usepackage{courier}
\usepackage{url}
\usepackage{graphicx}
\usepackage{amsmath}
\usepackage{amssymb}

\title{Code Completion with Neural Attention and Pointer Networks}

\author{Jian Li, Yue Wang, Michael R. Lyu, Irwin King\\
Department of Computer Science and Engineering, The Chinese University of Hong Kong, China\\
Shenzhen Research Institute, The Chinese University of Hong Kong, China\\
\{jianli, yuewang, lyu, king\}@cse.cuhk.edu.hk\\}

\begin{document}

\maketitle

\begin{abstract}
Intelligent code completion has become an essential research task to accelerate modern software development. To facilitate effective code completion for dynamically-typed programming languages, we apply neural language models by learning from large codebases, and develop a tailored attention mechanism for code completion. However, standard neural language models even with attention mechanism cannot correctly predict the out-of-vocabulary (OoV) words that restrict the code completion performance. In this paper, inspired by the prevalence of locally repeated terms in program source code, and the recently proposed pointer copy mechanism, we propose a pointer mixture network for better predicting OoV words in code completion. Based on the context, the pointer mixture network learns to either generate a within-vocabulary word through an RNN component, or regenerate an OoV word from local context through a pointer component. Experiments on two benchmarked datasets demonstrate the effectiveness of our attention mechanism and pointer mixture network on the code completion task.
\end{abstract}

\section{Introduction}
Integrated development environments (IDEs) have become essential paradigms for modern software engineers, as IDEs provide a set of helpful services to accelerate software development. Intelligent code completion is one of the most useful features in IDEs, which suggests next probable code tokens, such as method calls or object fields, based on existing code in the context. Traditionally, code completion relies heavily on compile-time type information to predict next tokens \cite{tu2014localness}. Thus, it works well for statically-typed languages such as Java. Yet code completion is harder and less supported for dynamically-typed languages like JavaScript and Python, due to the lack of type annotations.

To render effective code completion for dynamically-typed languages, recently, researchers turn to learning-based language models \cite{hindle2012naturalness,white2015toward,bielik2016phog}. They treat programming languages as natural languages, and train code completion systems by learning from large codebases (e.g., GitHub). In particular, neural language models such as Recurrent Neural Networks (RNNs) can capture sequential distributions and deep semantics, hence become very popular. However, these standard neural language models are limited by the so-called \emph{hidden state bottleneck}: all the information about current sequence is compressed into a fixed-size vector. The limitation makes it hard for RNNs to deal with long-range dependencies, which are common in program source code such as a class identifier declared many lines before it is used.

\begin{figure*}[ht]
  \centering
  \includegraphics[width=0.9\linewidth]{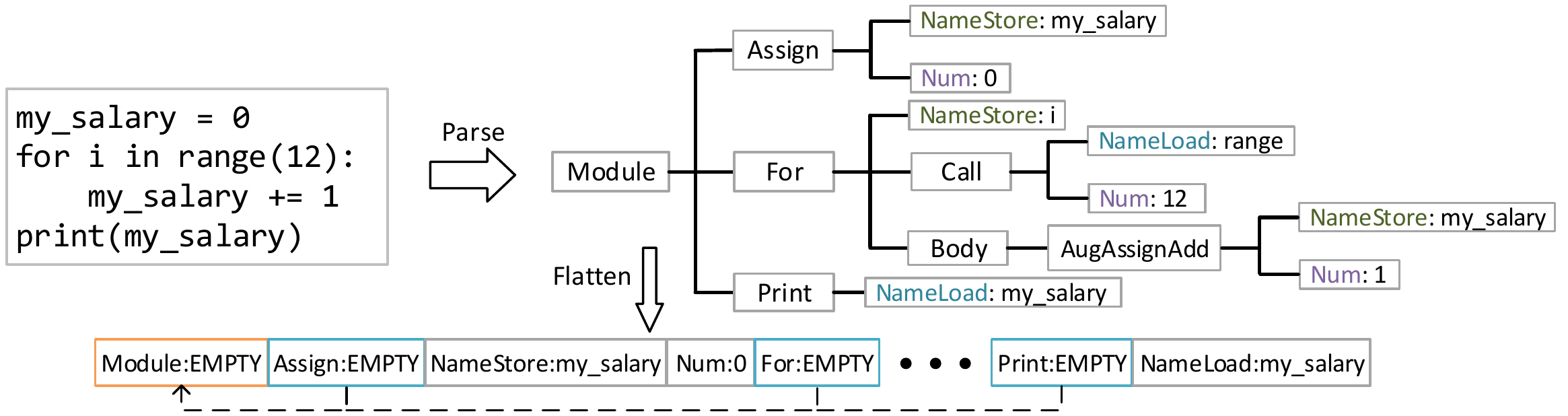}\\\vspace{-1ex}
  \caption{A Python program and its corresponding abstract syntax tree (AST). The dashed arrow points to a parent node.}\label{AST}\vspace{-2ex}
\end{figure*}

Attention mechanism \cite{bahdanau2014neural} provides one solution to this challenge. With attention, neural language models learn to retrieve and make use of relevant previous hidden states, thereby increasing the model's memorization capability and providing more paths for back-propagation. To deal with long-range dependencies in code completion, we develop a tailored attention mechanism which can exploit the structure information on program's \emph{abstract syntax tree} (AST, see Figure \ref{AST}), which will further be described.

But even with attention, there is another critical issue called \emph{unknown word problem}. In general, the last component of neural language models is a softmax classifier, with each output dimension corresponding to a unique word in the predefined vocabulary. As computing high-dimensional softmax is computational expensive, a common practice is to build the vocabulary with only \emph{K} most frequent words in the corpus and replace other out-of-vocabulary (OoV) words with a special word, i.e., \emph{UNK}. Intuitively, standard softmax based neural language models cannot correctly predict OoV words. In code completion, simply recommending an \emph{UNK} token offers no help to the developers. The \emph{unknown word problem} restricts the performance of neural language models, especially when there are a large number of unique words in the corpus like program source code.

For our code completion task, we observe that when writing programs, developers tend to repeat locally. For example, the variable name \texttt{\small{my\_salary}} in Figure \ref{AST} may be rare and marked as \emph{UNK} with respect to the whole corpus. But within that specific code block, it repeats several times and has a relatively high frequency. Intuitively, when predicting such unknown words, our model can \emph{learn to choose} one location in local context and copy the word at that location as our prediction. Actually, the recently proposed Pointer Networks \cite{vinyals2015pointer} can do so, which employ attention scores to select a word from the input sequence as output. Although pointer networks can make better predictions on unknown words or rare words, they are unable to predict words beyond current input sequence, i.e., lacking the global view. Therefore they may not work well in our code completion.

In this paper, to facilitate effective code completion, we propose a \emph{pointer mixture network}, which can predict next word by either generating one from the global vocabulary or copying a word from the local context. For the former, we apply a standard RNN with attention, which we call the \emph{global RNN component}. For the latter, we employ a pointer network which we call the \emph{local pointer component}. Actually the two components share the same RNN architecture and attention scores. Our pointer mixture network is a weighted combination of the two components. At each prediction, a switcher is learned based on the context information, which can guide the model to choose one component for generating the next word. In this way, our model learns \emph{when and where} to copy an OoV word from the local context as the final prediction.

The main contributions of this work are as follows:
\begin{itemize}
\item We propose a pointer mixture network for better predicting OoV words in code completion, which learns to generate next word from either the global vocabulary or the local context.
\item We develop an attention mechanism for code completion, which makes use of the AST structure information (specially, the parent-children information).
\item We evaluate our models on two benchmarked datasets (JavaScript and Python). The experimental results show great improvements upon the state-of-the-arts.
\end{itemize}
\vspace{-2ex}
\section{Approach}\label{approach}

\subsection{Program Representation}\label{program_represent}
In our corpus, each program is represented in the form of \emph{abstract syntax tree} (AST). Any programming language has an unambiguous context-free grammar, which can be used to parse source code into an AST. Further, the AST can be converted back into source code in a one-to-one correspondence. Processing programs in the form of ASTs is a typical practice in \emph{Software Engineering} (SE) \cite{mou2016convolutional,li2017DPCNN}.

Figure \ref{AST} shows an example Python program and its corresponding AST. We can see that each AST node contains two attributes: the \emph{type} of the node and an optional \emph{value}. For each leaf node, ``:'' is used as a separator between type and value. For each non-leaf node, we append a special \texttt{\small{EMPTY}} token as its value. As an example, consider the AST node \texttt{\small{NameLoad:my\_salary}} in Figure \ref{AST} where \texttt{\small{NameLoad}} denotes the type and \texttt{\small{my\_salary}} is the value. The number of unique types is relative small (hundreds in our corpus), with types encoding the program structure, e.g., \texttt{\small{Identifier,IfStatement, SwitchStatement}}, etc. Whereas there are infinite possibilities for values, which encode the program text. A value may be any program identifier (e.g. \texttt{\small{jQuery}}), literal (e.g. \texttt{\small{66}}), program operator (e.g., \texttt{\small{+,-,*}}), etc.

Representing programs as ASTs rather than plain text enables us to predict the structure of the program, i.e., type of each AST node. See the example in Figure \ref{AST} again, when the next token is a keyword \texttt{\small{for}}, the corresponding next AST node is \texttt{\small{For(:EMPTY)}}, which corresponds to the following code block:
\vspace{-0.5ex}
\begin{verbatim}
for __ in __:
  ## for loop body
\end{verbatim}
\vspace{-0.5ex}
In this way, successfully predicting next AST node completes not only the next token \texttt{\small{for}}, but also the whole code block including some trivial tokens like \texttt{\small{in}} and ``\texttt{\small{:}}''. Such structure completion enables more flexible code completion at different levels of granularity.

To apply statistical sequence models, we flatten each AST as a sequence of nodes in the in-order depth-first traversal. To make sure the sequence can be converted back to the original tree structure thus converted back to the source code, we allow each node type to encode two additional bits of information about whether the AST node has a child and/or a right sibling. If we define a word as $w_i=(T_i, V_i)$ to represent an AST node, with $T_i$ being the \emph{type} and $V_i$ being the \emph{value}, then each program can be denoted as a sequence of words $w_{i=1}^n$. Thus our code completion problem is defined as: given a sequence of words $w_1, ..., w_{t-1}$, our task is to predict the next word $w_t$. Obviously, we have two kinds of tasks: predicting the next node \emph{type} $T_t$ and predicting the next node \emph{value} $V_t$. We build one model for each task and train them separately. We call this AST-based code completion.

\subsection{Neural Language Model}
The code completion task can be regarded as a language modeling problem, where recurrent neural networks (RNNs) have achieved appealing success in recent years. LSTM \cite{hochreiter1997long} is proposed to mitigate the gradients vanishing/exploding problem in RNNs, by utilizing gating mechanisms. A standard LSTM cell is defined as $h_t=f(x_t, h_{t-1})$. At each time step $t$, an LSTM cell takes current input vector $x_t$ and previous hidden state $h_{t-1}$ as inputs, then produces the current hidden state $h_t$ which will be used to compute the prediction at time step $t$.

\begin{figure}[t]
  \centering
  \includegraphics[width=\linewidth]{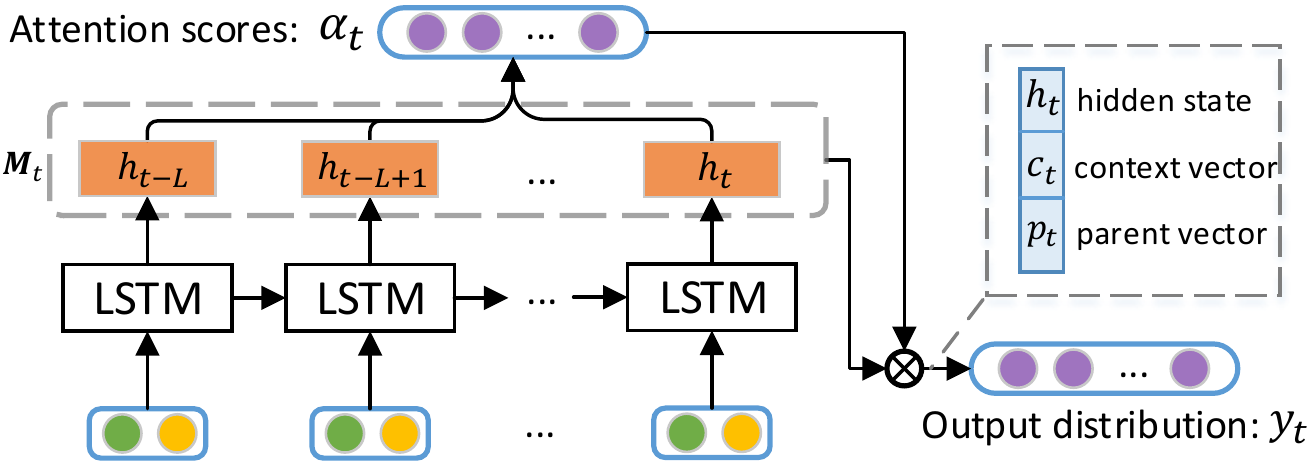}\vspace{-1ex}
  \caption{The attentional LSTM. The inputs fed to each LSTM cell are composed of two kinds of embeddings (green for Type and yellow for Value). Here $\otimes$ represents the element-wise multiplication.}\label{attention_figure}\vspace{-2ex}
\end{figure}

\subsection{Attention Mechanism}
Standard neural language models suffer from hidden state bottleneck \cite{cheng2016long}. To alleviate the problem, attention mechanism is proposed to retrieve and make use of \emph{relevant} previous hidden states. It is incorporated into standard LSTM which we call attentional LSTM in this paper, as illustrated in Figure \ref{attention_figure}.

\noindent\textbf{Context Attention} ~Traditional attention mechanism makes use of previous hidden states within a context window \cite{bahdanau2014neural}, which we call the \emph{context attention}. Formally, we keep an external memory of $L$ previous hidden states, which is denoted as $M_t=[h_{t-L}, ... , h_{t-1}]\in \mathbb{R}^{k*L}$. At time step $t$, the model uses an attention layer to compute the relation between $h_t$ and hidden states in $M_t$, represented as attention scores $\alpha_t$, and then produces a summary context vector $c_t$. We design our context attention for code completion as follows:
\begin{align} 
A_t &=  v^T \tanh(W^mM_t +(W^hh_t)1^T_L), \\ 
\alpha_t &=  softmax(A_t) \label{attention},\\ 
c_t &= M_t \alpha^T_t ,
\end{align}
where $W^m, W^h \in \mathbb{R}^{k*k}$ and $v \in \mathbb{R}^k$ are trainable parameters. $k$ is the size of the hidden state, i.e. dimension of $h_t$. $1_L$ represents an L-dimensional vector of ones.

\noindent\textbf{Parent Attention} ~Besides the traditional context attention, we also propose a \emph{parent attention} for the AST-based code completion. Intuitively, different hidden states within the context window should have different degrees of relevance to the current prediction. As our sequence is flattened from a tree (i.e., AST, see Figure \ref{AST}), a parent node should be of great relevance to a child node. But the flattened AST has lost the parent-children information. To exploit such structure information, when flattening the AST, we record the parent location $pl$ of each AST node, i.e., how many nodes before it. Then at time step $t$, our model retrieves a parent vector $p_t$ from the external memory $M_t$, which is the hidden state at the parent location, i.e., $h_{t-pl}$ \footnote{If $pl$ is larger than $L$, we set $pl$ as $1$.}. The information of parent code segments can benefit our model to make more confident predictions. 

When predicting next word at time step $t$, we condition the decision on not only the current hidden state $h_t$ but also the context vector $c_t$ and parent vector $p_t$. The output vector $G_t$ encodes the information about next token which is then projected into the vocabulary space, followed by a softmax function to produce the final probability distribution $y_t \in \mathbb{R}^{V}$:
\begin{align} 
G_t &=  \tanh(W^g[h_t;c_t;p_t]), \label{context and parent}\\ 
y_t &= softmax(W^vG_t + b^v), \label{vocab_softmax}
\end{align}
where $W^g \in \mathbb{R}^{k*3k}$ and $W^v \in \mathbb{R}^{V*k}$ are two trainable projection matrices and $b^v \in \mathbb{R}^{V}$ is a trainable bias vector. Note that $V$ represents the size of vocabulary and ``;'' denotes the concatenation operation.

\begin{figure}[t]
  \centering
  \includegraphics[width=\linewidth]{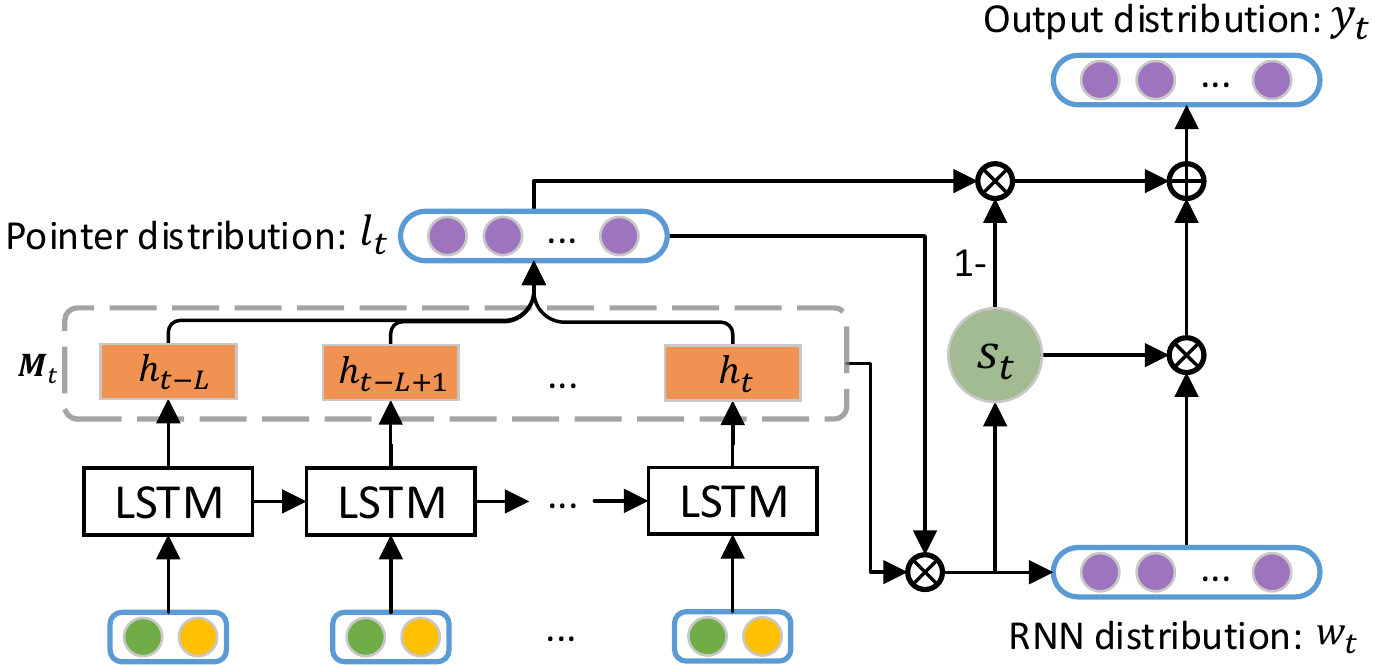}\vspace{-1ex}
  \caption{The pointer mixture network. We reuse the attention scores $\alpha_t$ (see Figure \ref{attention_figure}) as the pointer distribution $l_t$. The switcher produces $s_t\in[0,1]$ to balance $l_t$ and $w_t$. The final distribution is generated by concatenating the two scaled distributions. Here $\oplus$ indicates the concatenation operation.}\label{pointer_figure}\vspace{-2ex}
\end{figure}

\begin{table*}[ht]
\centering
\caption{Accuracies on next \emph{value} prediction with different vocabulary sizes. The out-of-vocabulary (OoV) rate denotes the percentage of AST nodes whose value is beyond the global vocabulary. Localness is the percentage of values who are OoV but occur in the context window.}\vspace{-1ex}
\label{RQ1}
\begin{tabular}{ccccccc}
\hline
Vocabulary Size & JS\_1k         & JS\_10k  & JS\_50k  & PY\_1k & PY\_10k  & PY\_50k       \\
OoV Rate / Localness & 20\% / 8\%   & 11\% / 3.7\%  & 7\% / 2\%  & 24\% / 9.3\% & 16\% / 5.2\%  & 11\% / 3.2\%  \\ \hline\hline

Vanilla LSTM   & 69.9\%          & 75.8\%          & 78.6\%          & 63.6\%          & 66.3\%          & 67.3\%       \\
Attentional LSTM (ours)  & 71.7\%       & 78.1\%       & 80.6\%       & 64.9\%       & 68.4\%       & 69.8\%      \\ Pointer Mixture Network (ours)    & \textbf{73.2\%} & \textbf{78.9\%} & \textbf{81.0\%} & \textbf{66.4\%} & \textbf{68.9\%} & \textbf{70.1\%} \\ \hline
\end{tabular}
\end{table*}

\subsection{Pointer Mixture Network}
Inspired by the prevalence of locally repeated tokens in program source code, we propose to leverage the pointer networks to predict OoV tokens in code completion, by \emph{copying} a token from previous input sequence. Specifically, we propose a \emph{pointer mixture network} that combines a standard RNN and a pointer network, as shown in Figure~\ref{pointer_figure}.

Our pointer mixture network consists of two major components (global RNN component and local pointer component), and one switcher to strike a balance between them. For global RNN component, it is an attentional LSTM that predicts the next token from a predefined global vocabulary. For local pointer component, it points to previous locations in local context according to the learned location weights. Our pointer mixture network combines the two components by concatenating the two components' output vectors. Before concatenation, the two individual outputs are scaled by a learned switcher based on the context, thus our model learns how to \emph{choose} a certain component at each prediction. Specifically, the switcher produces a scalar $s_t\in[0,1]$ which indicates the probability to use the global RNN component, and then $1-s_t$ is the probability to use the local pointer component.

After concatenating the two scaled vectors, we pick one output dimension with the highest probability. If this dimension belongs to the RNN component, then the next token is generated from the global vocabulary. Otherwise, the next token is copied from the local context.

Formally, at time step $t$, the global RNN component produces a probability distribution $w_t\in \mathbb{R}^{V}$ for the next token $x_t$ within the vocabulary according to formula \ref{vocab_softmax}. The local pointer component points to the locations inside a memory according to the distribution $l_t \in \mathbb{R}^{L}$, where $L$ is the length of the memory. In order to reduce the parameters and accelerate the training, we reuse the attention scores (see formula \ref{attention}) as $l_t$ in practice.

The switcher is a sigmoid function conditioned on the current hidden state $h_t$ and context vector $c_t$:
\begin{equation}
s_t=\sigma(W^s[h_t;c_t] + b^s),
\end{equation}
where $W^s \in \mathbb{R}^{2k*1}$ and $b^s \in \mathbb{R}^{1}$ are trainable weights. $s_t \in [0,1]$ is a scalar to balance $w_t$ and $l_t$. Finally, the model completes by concatenating the two scaled distributions to produce the final prediction:
\begin{equation}
y_t=[s_tw_t;(1-s_t)l_t].
\end{equation}

\section{Evaluation}
\subsection{Dataset}
We evaluate different approaches on two benchmarked datasets: JavaScript (JS) and Python (PY), which are summarized in Table \ref{dataset}. Collected from GitHub, both two datasets are publicly available\footnote{http://plml.ethz.ch} and used in previous work \cite{bielik2016phog,raychev2016probabilistic,liu2016neural}. Both datasets contain 150,000 program files which are stored in their corresponding AST formats, with the first 100,000 used for training and the remaining 50,000 used for testing. After serializing each AST in the in-order depth-first traversal, we generate multiple \emph{queries} used for training and evaluation, one per AST node, by removing the node (plus all the nodes to the right) from the sequence and then attempting to predict the node.

The numbers of unique node \emph{types} in JS and PY are 44 and 181 originally. By adding information about children and siblings as discussed in Section \ref{program_represent}, we increase the numbers to 95 and 330 respectively. As shown in Table \ref{dataset}, the number of unique node \emph{values} in both datasets are too large to directly apply neural language models, thus we only choose \emph{K} most frequent values in each training set to build the global vocabulary, where \emph{K} is a free parameter. We further add three special values: \texttt{\small{UNK}} for out-of-vocabulary values, \texttt{\small{EOF}} indicating the end of each program, and \texttt{\small{EMPTY}} being the value of non-leaf AST nodes.

\begin{table}[t]
\caption{Dataset Statistics \label{dataset}}\vspace{-1ex}
\begin{centering}
\begin{tabular}{c c c}
\hline
 & JS & PY \tabularnewline
\hline
Training Queries & $10.7*10^7$ & $6.2*10^7$ \tabularnewline
Test Queries & $5.3*10^7$ & $3.0*10^7$ \tabularnewline
Type Vocabulary & 95 &330 \tabularnewline
Value Vocabulary & $2.6*10^6$ & $3.4*10^6$ \tabularnewline
\hline
\end{tabular}
\par\end{centering}
\end{table}

\begin{table*}[ht]
\centering
\caption{Comparisons against the state-of-the-arts. The upper part is the results from our experiments while the lower part is the results from the prior work. TYPE means next node type prediction and VALUE means next node value prediction.}\vspace{-1ex}
\label{RQ2}
\begin{tabular}{ccccc}
\hline
                                              & \multicolumn{2}{c}{\textbf{JS}}   & \multicolumn{2}{c}{\textbf{PY}}   \\
                                              & TYPE            & VALUE           & TYPE            & VALUE           \\ \hline
\multicolumn{1}{c}{Vanilla LSTM}             & 87.1\%          & 78.6\%          & 79.3\%          & 67.3\%          \\
\multicolumn{1}{c}{Attentional LSTM (no parent attention)}  & 88.1\% & 80.5\%   & 80.2\% & 69.8\%          \\
\multicolumn{1}{c}{Attentional LSTM (ours)}  & \textbf{88.6\%} & 80.6\%          & \textbf{80.6\%} & 69.8\%          \\
\multicolumn{1}{c}{Pointer Mixture Network (ours)} & -          & 81.0\%          & -               & \textbf{70.1\%} \\ \hline

\multicolumn{1}{c}{LSTM \cite{liu2016neural}}                & 84.8\%     & 76.6\%     & -          & -          \\
\multicolumn{1}{c}{Probabilistic Model \cite{raychev2016probabilistic}}   & 83.9\%   & \textbf{82.9\%} & 76.3\%   & 69.2\%  \\ \hline
\end{tabular}
\end{table*}

\subsection{Experimental Setup}\label{label_UNK}
\noindent\textbf{Configuration} ~Our base model is a single layer LSTM network with unrolling length of 50 and hidden unit size of 1500. To train the model, we use the cross entropy loss function and mini-batch SGD with the Adam optimizer \cite{kingma2014adam}. We set the initial learning rate as 0.001 and decay it by multiplying 0.6 after every epoch. We clip the gradients' norm to 5 to prevent gradients exploding. The size of attention window is 50. The batch size is 128 and we train our model for 8 epochs. Each experiment is run for three times and the average result is reported.

We divide each program into segments consisting of 50 consecutive AST nodes, with the last segment being padded with \texttt{\small{EOF}} if it is not full. The LSTM hidden state and memory state are initialized with $h_0, c_0$, which are two trainable vectors. The last hidden and memory states from the previous LSTM segment are fed into the next one as initial states if both segments belong to the same program. Otherwise, the hidden and memory states are reset to $h_0, c_0$. We initialize $h_0, c_0$ to be all-zero vectors while all other variables are randomly initialized using a uniform distribution over [-0.05, 0.05]. We employ \emph{accuracy} as our evaluation metric, i.e., the proportion of correctly predicted next node types/values.

\noindent\textbf{Preprocessing and Training Details} ~As each AST node consists of a type and a value, to encode the node and input it to the LSTM, we train an embedding vector for each type (300 dimensions) and value (1200 dimensions) respectively, then concatenate the two embeddings into one vector. Since the number of unique types is relatively small in both datasets, there is no \emph{unknown word problem} when predicting next AST node type. Therefore, we only apply our \emph{pointer mixture network} on predicting next AST node value.

For each dataset, we build the global vocabulary for AST node values with \emph{K} most frequent values in the training set, and mark all out-of-vocabulary node values in training set and test set as OoV values. Before training, if an OoV value appears exactly the same as another previous value within the attention window, then we label that OoV value as the corresponding \emph{position} in the attention window. Otherwise, the OoV value is labeled as \texttt{\small{UNK}}. If there are multiple matches in the attention window, we choose the position label as the last occurrence of the matching value in the window, which is the closest one. For within-vocabulary values, we label them as the corresponding IDs in the global vocabulary. During training, whenever the ground truth of a training query is \texttt{\small{UNK}}, we set the loss function to zero for that query such that our model does not learn to predict \texttt{\small{UNK}}. In both training and evaluation, all predictions where the target value is \texttt{\small{UNK}} are treated as \emph{wrong} predictions, i.e., decrease the overall accuracy.


\subsection{Experimental Results}
For each experiment, we run the following models for comparison, which have been introduced in Section \ref{approach}:
\begin{itemize}
\item \textbf{Vanilla LSTM}: A standard LSTM network without any attention or pointer mechanisms.
\item \textbf{Attentional LSTM}: An LSTM network equipped with our (context and parent) attention mechanism which attends to last 50 hidden states at each time step.
\item \textbf{Pointer Mixture Network}: Our proposed mixture network which combines the above attentional LSTM and the pointer network.
\end{itemize}

\noindent\textbf{OoV Prediction} ~We first evaluate our pointer mixture network's ability to ease the \emph{unknown word problem} when predicting next AST node value. For each of the two datasets, we create three specific datasets by varying the global vocabulary size \emph{K} for node values to be 1k, 10k, and 50k, resulting in different out-of-vocabulary (OoV) rates. We also measure how often OoV values can occur in previous context window thus be labeled as the corresponding positions. We call this measure as \emph{localness}, which is the upper-bound of the performance gain we can expect from the pointer component. We run the above models on each specific dataset. Table \ref{RQ1} lists the corresponding statistics and experimental results.

As Table \ref{RQ1} shows in the column, on each specific dataset, the vanilla LSTM achieves the lowest accuracy, while the attentional LSTM improves the performance upon the vanilla LSTM, and our pointer mixture network achieves the highest accuracy. Besides, we can see that by increasing the vocabulary size in JS or PY dataset, the OoV rate decreases, and the general accuracies on different models increase due to more available information. We also notice a performance gain by our pointer mixture network over the attentional LSTM, and the gain is the largest with 1k vocabulary size. We attribute this performance gain to correctly predicting some OoV values through the local pointer component. Therefore, the results demonstrate the effectiveness of our pointer mixture network to predict OoV values, especially when the vocabulary is small and the OoV rate is large.

\begin{figure*}[ht]
  \centering
  \includegraphics[width=\linewidth]{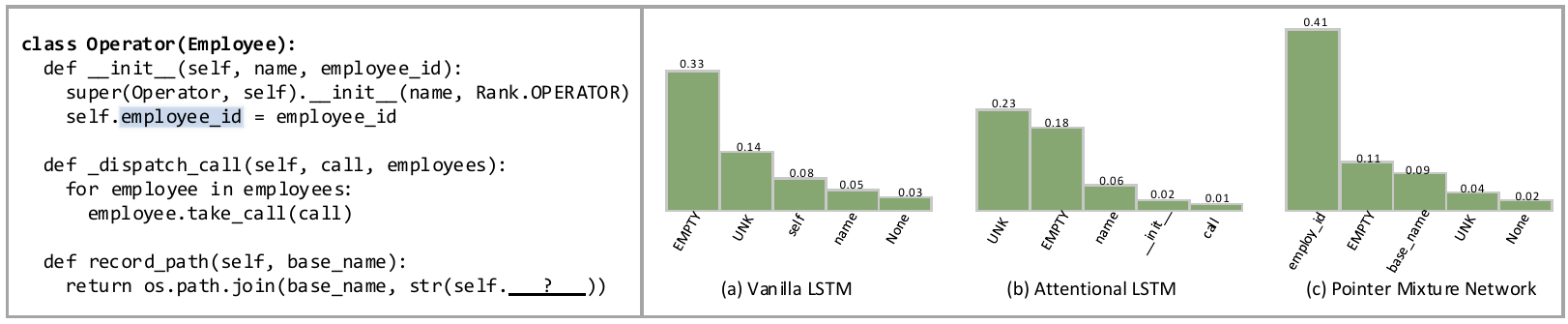}\vspace{-1ex}
  \caption{A code completion example showing predicting an OoV value.}\label{case}\vspace{-2ex}
\end{figure*}

\noindent\textbf{State-of-the-Art Comparison} ~As there are already prior investigations conducting code completion on the two benchmarked datasets, to validate the effectiveness of our proposed approaches, we need to compare them against the state-of-the-art. Particularly, Liu et al. \shortcite{liu2016neural} employ a standard LSTM on the JS dataset, without attention or pointer mechanisms. Raychev et al. \shortcite{raychev2016probabilistic} build a probabilistic model for code based on probabilistic grammars and achieve the state-of-the-art accuracies for code completion on the two datasets.

Specifically, we conduct experiments on next AST node type prediction and next AST node value prediction respectively. For the former, there is no \emph{unknown word problem} due to the small type vocabulary, so we only use the vanilla LSTM and the attentional LSTM. For the latter, we set the value vocabulary size to 50k to make the results comparable with \cite{liu2016neural}, and employ all the three models. The results are shown in Table \ref{RQ2}.

The upper part of Table \ref{RQ2} shows our results in this work, while the lower part lists the results from the prior work. Note that Liu et al. \shortcite{liu2016neural} only apply LSTM on the JS dataset, so they do not have results on the PY dataset. For next type prediction, our attentional LSTM achieves the highest accuracy on both datasets, significantly improving the best records of the two datasets. For next value prediction on JS dataset, our pointer mixture network achieves comparable performance with Raychev et al.'s \shortcite{raychev2016probabilistic}, which is a probabilistic model based on domain-specific grammars. However, our approaches outperform Liu et al. \shortcite{liu2016neural} that is also based on neural networks. On PY dataset, our pointer mixture network for next value prediction outperforms the previous best record. Therefore, we conclude that our attentional LSTM and pointer mixture network are effective for code completion, achieving three state-of-the-art performances out of the four tasks.

\subsection{Discussion}

\noindent\textbf{Why attention mechanism works?} ~When writing programs, it is quite common to refer to a variable identifier declared many lines before. In this work, the mean program length (i.e., the number of AST nodes) is around 1000 in JS dataset and 600 in PY dataset. Therefore in our code completion task, we need the attention mechanism to capture the long dependencies. Furthermore, we measure how our proposed \emph{parent attention} influence the final prediction by only using the context attention (see formula \ref{context and parent}). As shown in Table \ref{RQ2}, parent attention can effectively contribute to the type prediction while has little effect on the value prediction.

\begin{table}[t]
\centering
\caption{Showing why pointer mixture network works.}\vspace{-1ex}
\label{RQ3}
\begin{tabular}{lll}
\hline
                        & JS\_1k & PY\_1k \\ \hline
Pointer Random Network  & 71.4\% & 64.8\% \\
Attentional LSTM & 71.7\% & 64.9\% \\
Pointer Mixture Network & \textbf{73.2}\% & \textbf{66.4}\% \\ \hline
\end{tabular}
\end{table}

\noindent\textbf{Why pointer mixture network works?} ~After incorporating the pointer network, we predict OoV values by copying a value from local context and that copied value may be the correct prediction. Thus we observe a performance gain in our pointer mixture network. However, one may argue that no matter how capable the pointer component is, the accuracy will definitely increase as long as we get chances to predict OoV values.

To verify the copy ability of our pointer component, we develop a \emph{pointer random network} where the pointer distribution $l_t$ (see Figure \ref{pointer_figure}) is a random distribution instead of reusing the learned attention scores. We conduct comparisons on value prediction in JS and PY datasets with 1k vocabulary size. The results are listed in Table \ref{RQ3}, where the pointer random network achieves lower accuracies than the pointer mixture network. Thus we demonstrate that our pointer mixture network indeed learns \emph{when and where} to copy some OoV values. However, the pointer random network performs even worse than the attentional LSTM. We think the reason lies in the switcher which is disturbed by the random noise and cannot always choose the correct component (i.e., the RNN component), thus influencing the overall performance. 

\vspace{-0.5ex}
\subsection{Case Study}
We depict a code completion example in Figure \ref{case}. In this example, the target prediction \texttt{\small{employee\_id}} is an OoV value with respect to the whole training corpus. We show the top five predictions of each model. For vanilla LSTM, it just produces \texttt{\small{EMPTY}} which is the most frequent node value in our corpus. For attentional LSTM, it learns from the context that the target has a large probability to be \texttt{\small{UNK}}, but fails to produce the real value. Pointer mixture network successfully identifies the OoV value from the context, as it observes the value appearing in the previous code.

\vspace{-0.5ex}
\section{Related Work}

\noindent\textbf{Statistical Code Completion} ~There is a body of recent work that explores the application of statistical learning and sequence models on the code completion task, such as $n$-gram models \cite{hindle2012naturalness,tu2014localness}, and probabilistic grammars \cite{allamanis2014mining,bielik2016phog,raychev2016probabilistic}. Recently, neural networks become very popular to model source code \cite{raychev2014code,white2015toward,allamanis2016convolutional}. In particular, Bhoopchand et al. \shortcite{bhoopchand2016learning} proposed a sparse pointer mechanism for RNN, to better predict identifiers in Python source code. Nevertheless, their pointer component targets at identifiers in Python source code, rather than OoV tokens in our work. The OoV tokens include not only identifiers but also other types such as VariableDeclarator. Besides, they directly serialize each program as a sequence of code tokens, while in our corpus each program is represented as a sequence of AST nodes to facilitate more intelligent structure prediction.

\noindent\textbf{Neural Language Modeling} ~Deep learning techniques such as RNNs have achieved the state-of-the-art results in the language modeling task \cite{mikolov2010recurrent}. The soft attention or memory mechanisms \cite{bahdanau2014neural,cheng2016long,tran2016recurrent} have been proposed to ease the gradient vanishing problem in standard RNNs. Pointer is another mechanism proposed recently \cite{vinyals2015pointer} which gives RNNs the ability to ``copy''. The pointer mechanism is shown to be helpful in tasks like summarization \cite{gu2016incorporating}, neural machine translation \cite{luong2014addressing}, code generation \cite{ling2016latent}, and language modeling \cite{merity2016pointer}.

Specially, Gulcehre et al. \shortcite{gulcehre2016pointing} also propose to generate new words at each time step based on an RNN component and a local pointer component. However, their scenario is sequence-to-sequence tasks like neural machine translation while our scenario is language modeling. Merity et al. \shortcite{merity2016pointer} also share a similar idea. But they employ the pointer component to effectively reproduce rare words which are still IDs in the global vocabulary, rather than out-of-vocabulary words. Further, the two work's corpora are natural language while our corpus is program source code.

\vspace{-0.5ex}
\section{Conclusion}
In this paper, we apply neural language models on the code completion task, and develop an attention mechanism which exploits the parent-children information on program's AST. To deal with the OoV values in code completion, we propose a pointer mixture network which learns to either generate a new value through an RNN component, or copy an OoV value from local context through a pointer component. Experimental results demonstrate the effectiveness of our approaches.

\section*{Acknowledgment}
The work described in this paper was fully supported by the National Natural Science Foundation of China (Project No. 61332010 and No. 61472338), the Research Grants Council of the Hong Kong Special Administrative Region, China (No. CUHK 14234416 and 14208815 of the General Research Fund), and Microsoft Research Asia (2018 Microsoft Research Asia Collaborative Research Award).

\bibliographystyle{named}
\bibliography{ijcai18}

\end{document}